\documentclass[letterpaper]{article} 
\usepackage{times}  
\usepackage{helvet}  
\usepackage{courier}  
\usepackage[hyphens]{url}  
\usepackage{graphicx} 
\usepackage{natbib}  
\usepackage{caption} 
\DeclareCaptionStyle{ruled}{labelfont=normalfont,labelsep=colon,strut=off} 
\usepackage{epsfig}
\usepackage{amsmath}
\usepackage{amssymb, array}
\usepackage{algorithm,algpseudocode}
\usepackage{comment}

\graphicspath{{figures/}} 

\newcommand{\etal}{\textit{et al}.}

\newcommand{\eg}{\textit{e}.\textit{g}.}

\newlength\myindent
\newcolumntype{C}{>{\centering}p{0.02\textwidth}}

\title{Semi-supervised Domain Adaptation for Semantic Segmentation}
\author{Ying Chen, Xu Ouyang, Kaiyue Zhu, Gady Agam}

\begin{document}

\maketitle

\begin{abstract}
Deep learning approaches for semantic segmentation rely primarily on supervised learning approaches and require substantial efforts in producing pixel-level annotations. Further, such approaches may perform poorly when applied to unseen image domains. To cope with these limitations, both unsupervised domain adaptation (UDA) with full source supervision but without target supervision and semi-supervised learning (SSL) with partial supervision have been proposed. While such methods are effective at aligning different feature distributions, there is still a need to efficiently exploit unlabeled data to address the performance gap with respect to fully-supervised methods. In this paper we address semi-supervised domain adaptation (SSDA) for semantic segmentation, where a large amount of labeled source data as well as a small amount of labeled target data are available. We propose a novel and effective two-step semi-supervised dual-domain adaptation (SSDDA) approach to address both cross- and intra-domain gaps in semantic segmentation. The proposed framework is comprised of two mixing modules. First, we conduct a cross-domain adaptation via an image-level mixing strategy, which learns to align the distribution shift of features between the source data and target data. Second, intra-domain adaptation is achieved using a separate student-teacher network which is built to generate category-level data augmentation by mixing unlabeled target data in a way that respects predicted object boundaries. We demonstrate that the proposed approach outperforms state-of-the-art methods on two common synthetic-to-real semantic segmentation benchmarks. An extensive ablation study is provided to further validate the effectiveness of our approach.
\end{abstract}

\begin{figure*}[h]

\begin{minipage}[b]{1.0\linewidth}
  \centering
  \centerline{\includegraphics[scale=0.60]{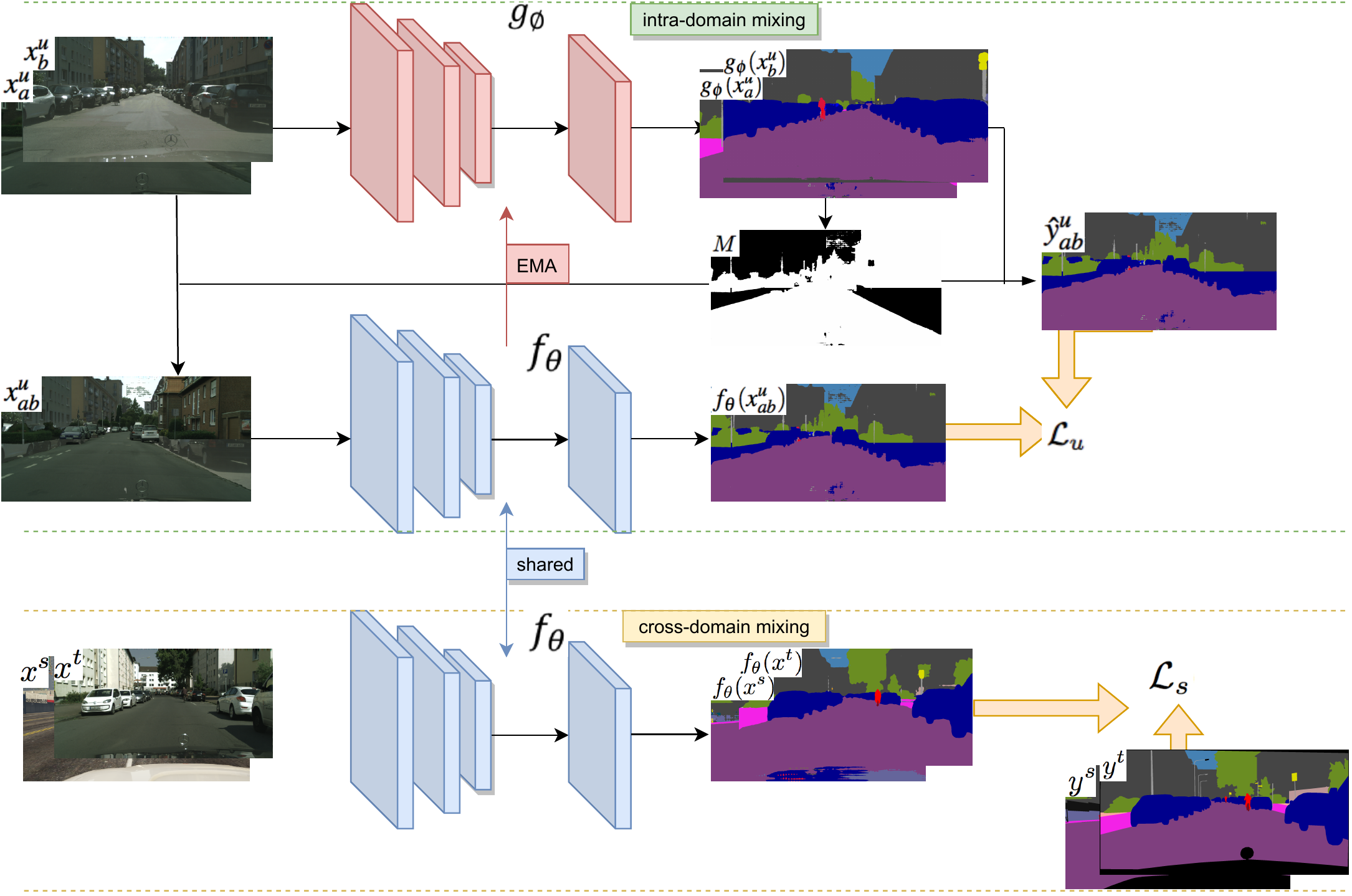}} 
\end{minipage}
\caption{Illustration of our proposed approach to semi-supervised dual-domain adaptation for semantic segmentation. The top and middle branches in this network comprise a student-teacher network which is trained on unlabeled target data to address intra-domain discrepancy within the target domain, whereas the bottom branch is a student network that shares weights with the middle student network in order to address cross-domain shift between labeled source and labeled target data. The student and teacher networks are linked using exponential moving average (EMA).}
\label{fig:arc}
\end{figure*}

\section{Introduction}
Semantic segmentation, which assigns semantic labels to each pixel in an image, is an essential tasks for many computer vision applications.
Deep neural networks show remarkable success in semantic segmentation (\eg~\cite{liu2019auto, takikawa2019gated}). However, the success of current deep semantic segmentation methods depend on having large quantities of manually annotated data which may be difficult to produce for dense prediction tasks. As shown in~\cite{feng2020semi}, the performance of a semantic segmentation network degrades rapidly with limited annotations even with carefully tuned networks. Further, data-driven approaches with limited data may not generalize well outside the training environment.

Unsupervised domain adaptation (UDA) attempt to mitigate the burden of annotation by using large amounts of synthetic data generated with known labels (\eg using a graphics engine) as a source domain (\eg GTA5~\cite{richter2016playing} and SYNTHIA~\cite{ros2016synthia}) and unlabeled examples of a target domain. The goal of UDA is to transfer learned knowledge from a labeled source domain to an unlabeled target domain by bridging the domain shift between them. Recent UDA methods have achieved significant improvements by employing different methods such as self-training~\cite{zou2018unsupervised, zou2019confidence, li2019bidirectional, zhang2019category, zhang2021prototypical}, generative models~\cite{hoffman2018cycada, kim2020learning}, adversarial learning~\cite{tsai2018learning, vu2019advent} and entropy minimization~\cite{ vu2019dada, pan2020unsupervised}. However, due to not fully capturing information from target domain data, there is still a large performance gap between such solutions and their fully-supervised counterparts.

Semi-supervised learning (SSL) is an another approach that aims at mitigating the burden of annotation. In such approaches a small set of labeled target data is used to label the remaining unlabeled data in an automated way thus leveraging the unlabeled data. SSL has shown significant progress for semantic segmentation in recent years~\cite{feng2020semi, french2019semi, mittal2019semi, ouali2020semi, olsson2021classmix}. A potential problem with such approaches is overfitting of learned models to the small set of labeled target data, hence leading to a need to improve model generalization.
Approaches for SSL include GAN-based adversarial learning~\cite{hung2018adversarial, mittal2019semi},  
pseudo supervision~\cite{feng2020semi,chen2020leveraging}, and consistency regularization~\cite{french2019semi,kim2020structured,olsson2020classmix}.




%
Semi-supervised domain adaptation (SSDA) combines UDA and SSL by using a large set of labeled source data, a small set of labeled target data, and a large set of unlabeled target data. SSDA has been applied to image classification~\cite{kim2020attract, qin2020opposite, rukhovich2019mixmatch, yang2020mico} and semantic segmentation \cite{wang2020alleviating, chen2021semi}. Wang \etal~\cite{wang2020alleviating} propose a semi-supervised approach termed "Alleviating Semantic-level Shift" (ASS) to promote distribution consistency between global and local views. 

SSDA shares challenges with both conventional UDA and SSL approaches. Although SSDA can be approached by directly using the labeled target data to supervise the segmentation task together with the supervised source data, doing so may lead to intra-domain discrepancy in the target domain which was demonstrated in semi-supervised domain adaptation for image classification task~\cite{kim2020attract} where the target domain is separated into two unaligned sub-distributions due to aligned labeled source and labeled target data. That is, enforcing partial alignment between full labeled source data and a few labeled target data does not guarantee that the remaining unlabeled target samples will be attracted toward source feature clusters, thus leaving them unaligned. The presence of the intra-domain discrepancy makes traditional domain adaptation methods less suitable for SSDA. 
To address SSDA, Chen \etal~\cite{chen2021semi} propose a two-step domain mixing model to effectively utilize labeled data in different domains by attempting to extract domain-invariant representations and so reduce the domain gap. In this approach the cross-domain region-level mixing employs CutMix~\cite{yun2019cutmix} and so fails to respect object boundaries. Furthermore, cross-domain mixing during domain adaptation doesn't take into account the intra-domain discrepancy within the target domain, which has never been introduced and tackled in SSDA based semantic segmentation. 
%

%

Following the intra-domain discrepancy observation described before, we propose a new semi-supervised dual-domain adaptation (SSDDA) approach for semantic segmentation to align cross-domain features by addressing the intra-domain discrepancy within the target domain.
The proposed approach overcomes the limitations of previous methods and 
improves the performance of deep classifiers on target domains with a small 
set of labeled targets.
This is achieved by applying our proposed mask-basked data augmentation strategy in two ways. First, by perturbing unlabeled target samples we attempt to capture unseen variations within the target domain to bridge unaligned target sub-distributions to some extent. Second, by employing a consistency loss over unlabeled mixed samples from the target domain we effectively alleviate the intra-domain gap within the target domain (due to cross-domain adaptation between source data and the labeled target data). 


Our proposed SSDDA approach utilizes a two-step domain adaptation: a cross-domain adaptation and an intra-domain adaptation. The goal of the cross-domain adaptation is to learn discriminative feature representation between source and target domains, while intra-domain adaptation aims at addressing intra-domain discrepancy within the target domain caused by the cross-domain adaptation. The intra-domain adaptation is performed by mixing two unlabeled target samples using their predicted segmentation maps. The cross-domain adaptation method is performed by mixing labeled images from two domains so as to extract domain-invariant representation.

\medskip\noindent
\textbf{Contributions.}
Our main contributions can be summarized as follows: (1) We address target intra-domain discrepancy in SSDA for semantic segmentation. (2) We propose a dual-domain adaptation framework that minimizes global domain shift between source and target domains while maintaining already aligned labeled and unlabeled target samples. (3) We demonstrate the effectiveness of our approach by improving state-of-the-art results on two common synthetic-to-real datasets.



\section{Related Work}

Domain adaptation aims at helping a trained model better generalize to unseen test data. Adversarial training~\cite{hoffman2016fcns, chen2018road, hong2018conditional, tsai2018learning, vu2019advent, chen2019domain} is a typical structure for UDA for semantic segmentation, which aims to address domain discrepancy between source and target domains by aligning feature distributions via an adversarial network. UDA via an adversarial network consists of two sub-networks. A discriminator aims to distinguish between domain the input feature maps, while a generator aims at generating feature maps to fool the discriminator. In this way, the discriminator is able to minimize the discrepancy of feature representations in the two domains. A similar approach for domain adaptation for semantic segmentation uses generative networks by conditioning target images on the source domain~\cite{hoffman2018cycada, wu2018dcan, li2019bidirectional}.  
Another category of approaches addressing UDA involves self-training~\cite{he2021multi, kim2020learning, li2019bidirectional, zhang2021prototypical}. The core idea in such approaches is to generate pseudo labels for unlabeled target domain data using an ensemble of previous models.
While, recent UDA methods for semantic segmentation have achieved significant progress, there is still a large performance gap in comparison to their fully supervised counterparts without target supervision.

A common way to avoid the need for a large labeled set for a dense prediction task such as semantic segmentation, is to annotate a small set of examples and use a semi-supervised learning approach for the remaining unlabeled data to train a model. Semi-supervised semantic segmentation has been widely studied using different mechanisms. Some methods utilize GAN-based adversarial learning~\cite{hung2018adversarial,mittal2019semi}, while others~\cite{feng2020semi,chen2020leveraging} generate pseudo supervision from the trained model (\eg self-training). 
Consistency regularization can be used to address Semi-supervised semantic segmentation~\cite{french2019semi,kim2020structured,olsson2020classmix}. This is done by enforcing a learned model to produce robust predictions for perturbations of unlabeled examples. Consistency regularization for semantic segmentation was first successfully used in a medical imaging application and has since been applied to other domains. Olsson~\cite{olsson2020classmix} propose a similar technique along this line by using predictions of a segmentation network to construct mixing, thus encouraging consistency over highly varied mixed samples while respecting the semantic boundaries of the original images. Our proposed method is inspired by~\cite{olsson2020classmix, chen2021mask} to make use of ideas from domain adaptation as well as semi-supervised learning.

Semi-supervised domain adaptation (SSDA) is an essential task that bridges the gap between a well-organized source domain and a target domain, thus promoting distribution consistency of features between the two domains using a small set of labeled data. Recent semi-supervised domain adaptation methods have shown remarkable progress for image classification~\cite{kim2020attract, qin2020opposite, rukhovich2019mixmatch, yang2020mico}. Kim \etal~\cite{kim2020attract} treat SSDA as two separate problems: UDA and SSL. They argue that conventional domain adaptation and semi-supervised learning methods result in less effective or negative transfer in SSDA, and so they propose to align features by alleviating intra-domain discrepancy. This is done by adopting co-training~\cite{chen2011co} to exchange the expertise between two classifiers which are trained on mixed data via a technique termed MixUp~\cite{zhang2017mixup} to mix labeled and unlabeled data in the target domain. The proposed approach addresses a similar issue in semantic segmentation. 


Several methods have been proposed to address SSDA in semantic segmentation using deep learning. Wang \etal~\cite{wang2020alleviating} propose a semi-supervised approach termed "Alleviating Semantic-level Shift" (ASS) to promote distribution consistency between global and local views. However, supervision of labeled target samples cannot fully take advantage of labeled data in the two domains. Further, the adversarial loss used in this approach makes the training unstable due to weak supervision. 
Chen \etal~\cite{chen2021semi} developed a framework based on a two-step domain adaptation technique that is realized using two complementary domain-mixed teachers from a holistic and partial views, and a student model for distilling knowledge from the two teachers. The cross-domain region-level mixing in this approach uses CutMix~\cite{yun2019cutmix} and so fails to respect object boundaries.
%
Intra-domain discrepancy in SSDA based semantic segmentation has not been addressed by these methods. Our proposed approach addresses these limitations and introduces a new perspective of intra-domain discrepancy within the target domain.


\section{Proposed Approach}


\medskip\noindent
\textbf{Problem Formulation}

\medskip\noindent
The method described in this paper is concerned with semi-supervised domain adaptation (SSDA) for semantic segmentation, given a large amount of labeled source data, a large amount of unlabeled target data, and a small amount of labeled target data. Denote the source domain data by $D_{s} = \{(x_{i}^{s}, y_{i}^{s})\}_{i=1}^{N_{s}}$, where $N_{s}$ is the number of labeled source examples and $x_{i}^{s}$ and $y_{i}^{s}$ represent an image example and its associated label respectively. Denote the labeled and unlabeled target domain data by $D_{t} = \{(x_{i}^{t}, y_{i}^{t})\}_{i=1}^{N_{t}}$ and $D_{u} = \{x_{i}^{u}\}_{i=1}^{N_{u}}$ respectively, where $N_{t}$ and $N_{u}$ is the number of labeled and unlabeled target examples. Our proposed SSDA approach aims to train an effective model on ($D_{s}$, $D_{t}$, $D_{u}$) with good generalization on unseen target examples by enhancing the target domain discriminability.

\medskip\noindent
\textbf{Proposed Dual-domain Adaptation}

\medskip\noindent
Domain shift between source and target data causes performance degradation. To address this problem we propose a two-step domain adaptation for SSDA. The proposed approach consists of a cross-domain adaptation network and an intra-domain adaptation and hence is termed dual-domain adaptation. The cross-domain part aims to close cross-domain shift, while the intra-domain part addresses intra-domain discrepancy generated by the cross-domain part. 
%
The cross-domain shift is addressed by training a single model on both labeled source data $D_{s}$ and labeled target data $D_{t}$. The intra-domain shift is addressed by utilizing unlabeled target data $D_{u}$. An overview of the proposed dual SSDA framework is shown in Figure~\ref{fig:arc}. Note that we only show the ComplexMix based framwork but don't show ClassMix based framework which is similar to ComplexMix based framework. Our framework consists of two main components: a student-teacher network for intra-domain adaptation with a teacher (upper branch) and a student (middle branch), and a second student network for cross-domain adaptation (lower branch) that shares weights with the the intra-domain student network (middle branch). The proposed approach follows a Mean Teacher Framework~\cite{tarvainen2017mean}, where the teacher network blends its parameters with the student network parameters using an exponential moving average (EMA).  Details of the main components are provided below.


%
The labeled examples from the source and target domains constitute domain examples with a large domain gap. So, we try to alleviate this cross-domain gap to enforce distribution consistency between the source and target domains by only mixing labeled images from both domains, and so by passing them through the same student network (middle branch in Figure~\ref{fig:arc}) we train the network to fit well labeled data from both domains. This process eliminates per-class decision boundaries lying between domains. With joint supervision of examples from both domains, the trained model tends to produce high-confidence predictions for intermediate features of both source-like and target-like images.

%
As described above, intra-domain discrepancy causes a distribution gap between sub-distributions of the target domain. In the context of SSDA intra-domain discrepancy occurs when the supervision of the samples from the source and target domains enforces the trained model to partially align source and target examples, thus leading to the presence of two unaligned sub-distributions corresponding to labeled and unlabeled data in the target domain. 
Following Olson \etal~\cite{olsson2020classmix, chen2021mask}, we propose an intra-domain mixing scheme using a student-teacher network to alleviate the intra-domain gap between sub-distributions within the target domain via category-level mixing that is applied to unlabeled data in the target domain. We feed an input image pair to the teacher network and a mixed image to the student network. We then enforce matching of blended input student predictions to blended teacher predictions.

\begin{table*}
\caption{Evaluation results for SYNTHIA$\rightarrow$Cityscapes. We follow previous works by training the model with $16$ classes and displaying the mIoU score of $13$ classes. 
The symbol $-$ indicates evaluations that cannot be performed.
The best results are highlighted.}
\begin{center}
\begin{tabular}{l|l|ccccc}
   \hline
  Category & Method & $0$ & $100$ & $200$  & $500$ & $1000$ \\
  \hline
  		&AdaptSeg~\cite{tsai2018learning}         &46.7  &- &- &- &-\\
  		&Advent~\cite{vu2019advent}      &48.0  &- &- &- &-\\
  UDA&LTI~\cite{kim2020learning} &49.3  &- &- &- &- \\
		&PIT~\cite{lv2020cross} &51.8 &- &- &- &-\\
		&ProDA~\cite{zhang2021prototypical} &62.0  &- &- &- &- \\ 
  \hline
  Supervised&DeeplabV2 &- &53.0 &58.9 &61.0 &67.5 \\  
  \hline  
  SSL&CutMix~\cite{french2019semi}      &-  &61.3 &66.7 &71.1 &73.0\\
  		&DST-CBC~\cite{feng2020semi}      &-  &59.7 &64.3 &68.9 &70.5\\
  \hline
  		& Baseline     &- &58.5 &61.9 &64.4 &67.6\\
  		& MME~\cite{saito2019semi}  &-  &59.6 &63.2 &66.7 &68.9\\
  SSDA&  ASS~\cite{wang2020alleviating} &-  &62.1 &64.8 &69.8 &73.0\\
  		&  DLDM~\cite{chen2021semi} &- &68.4 &69.8 &71.7 &74.2\\
  		&  Ours$^{1}$ (ClassMix)&-  &69.4 &71.6 &\textbf{73.2}&\textbf{74.7}\\
  		&  Ours$^{2}$ (ComplexMix)&-  &\textbf{70.6} &\textbf{71.8} &72.6&74.0\\
  \hline
\end{tabular}
\label{table:syn}
\end{center}
\end{table*}

\medskip\noindent
\textbf{Intra-domain Mixing}

\medskip\noindent
The intra-domain mixing we propose is illustrated in Figure~\ref{fig:arc}. We use a teacher (upper branch) and a student (middle branch). The intra-domain mixing we propose is implemented as follows: (1) Two unlabeled images $x_{a}^{u}$ and $x_{b}^{u}$ are randomly sampled from the unlabeled dataset; (2) The unlabeled samples $x_{a}^{u}$ and $x_{b}^{u}$ are fed into the teacher network which was trained on the labeled set to produce predictions $\hat{y}_{a}^{u}$ and $\hat{y}_{b}^{u}$; (3) We split $x_{a}^{u}$ and its corresponding semantic label $\hat{y}_{a}^{u}$ into $p\times p$ equal size blocks; (4) In each block, half of the classes are randomly selected using the argmaxed prediction of $x_{a}^{u}$ to generate a mask $M$. The binary mask $M$ contains $1$ values in chosen pixels and $0$ values otherwise; (5) The binary mask $M$ is used to create a mixed image $x_{ab}^{u}$ by using pixels from $x_{a}^{u}$ for mask values of $1$ and pixels from $x_{b}^{u}$ otherwise. The pseudo label for $x_{ab}^{u}$ based on the predictions of $x_{a}^{u}$ and $x_{b}^{u}$ is produced in a similar way; (6) Finally, the mixed samples $x_{ab}^{u}$ are fed into the student network to generate the mixed predictions. 

The mixing process as described above can be formulated using the following equations:
\begin{equation}
\begin{split}
x_{ab}^{u} = M \odot x_{a}^{u} + (1 - M) \odot x_{b}^{u} \\
\hat{y}_{ab}^{u} = M \odot \hat{y}_{a}^{u} + (1 - M) \odot \hat{y}_{b}^{u}
\end{split}
\end{equation}
where $\odot$ denotes element-wise multiplication.

The intra-domain image mixing we propose is able to produce new examples$x_{ab}^{u}$ that captures many kinds of unseen variations within the target domain. Such new examples bridge the gap between unaligned sub-distributions within the target domain to some extent. More importantly, we also employ a consistency loss~\ref{eq: seg3} to effectively alleviate the intra-target-domain gap generated due to cross-domain adaptation between the source and labeled target data. 
Specifically, we force agreement on target domain representations between the slow teacher network (whose parameters updated using moving average and hence is more stable and robust) and the fast student network. The motivation behind this approach is that, without any intra-domain regularization for the target domain, the cross-domain batch optimization would lean towards accumulating intra-domain differences. We thus pass the unlabeled data to both the slow teacher network (that is less suffering from intra-domain issues and is frozen) and the fast-updating student network (which potentially suffers from intra-domain gaps). Note that in this way if the intra-domain gap keeps growing, the consistency loss will increase, and so optimizing the consistency loss during training will help mitigate the generated intra-domain discrepancy within target domain. 


\newpage
\medskip\noindent
\textbf{Pseudo-labeling}

\medskip\noindent
In our intra-domain adaptation approach the label of the mixed image example is generated by directly taking argmax of the predictions produced by the teacher network $g_{\phi}$. Such labels are termed pseudo labels. The generated pseudo labels are used to enforce consistency over the mixed examples. In addition, pseudo-labeling is capable of eliminating uncertainty along boundaries. Semantic segmentation boundaries are the most difficult to predict and especially in an intra-domain mix where the context around the boundaries may be inaccurate. Pseudo-labeling address this issue by converting produced probabilities to one-hot encoded classification results.

\medskip\noindent
\textbf{Training}

\medskip\noindent
Our proposed model is trained to minimize a combined loss composed of two supervised loss terms $\mathcal{L}_{s}(*)$ and $\mathcal{L}_{t}(*)$, and an unsupervised consistency loss term $\mathcal{L}_{u}(*)$ designed to enforce consistency regularization between the student and teacher classifiers for unlabeled target images:
\begin{equation}
\mathcal{L}=\mathcal{L}_{s}(f_{\theta}(x^{s}),y^{s})+\mathcal{L}_{t}(f_{\theta}(x^{t}),y^{t})+\lambda\mathcal{L}_{u}(f_{\theta}(x^{u}),g_{\phi}(x^{u}))
\label{eq: loss}
\end{equation}
where $\lambda$ is a hyper-parameter used to control the balance between the supervised and unsupervised terms.

The student model $f_{\theta}$ is trained on labeled images of both the source and target domains in a supervised manner using the categorical cross entropy loss:
\begin{equation}
\mathcal{L}_{s}(f_{\theta}(x^{s}),y^{s})=-\dfrac{1}{N}\sum_{i=1}^{N}\sum_{j=1}^{H \times W}\sum_{c=1}^{C} y^{s}[i,j,c]\log f_{\theta}(x^{s})[i,j,c]
\label{eq: seg1}
\end{equation}
\begin{equation}
\mathcal{L}_{t}(f_{\theta}(x^{t}),y_{t})=-\dfrac{1}{N}\sum_{i=1}^{N}\sum_{j=1}^{H \times W}\sum_{c=1}^{C} y^{t}[i,j,c]\log f_{\theta}(x^{t})[i,j,c]
\label{eq: seg2}
\end{equation}
where $N$ is the total number of labeled examples. In this equation, ($f_{\theta}(x^{s})[i,j,c]$, $y^{s}[i,j,c]$) and ($f_{\theta}(x^{t})[i,j,c]$, $y^{t}[i,j,c]$) are the target and predicted probability pairs for pixel $(i, j)$ belonging to class $c$ for labeled source and target domains respectively.

The student model $f_{\theta}$ is also trained on unlabeled image pairs $u_{a}$ and $u_{b}$ using the categorical cross entropy loss as a consistency loss to match pseudo labels:
\begin{equation}
\mathcal{L}_{u}(f_{\theta}(x_{ab}^{u}),\hat{y}_{ab}^{u})=-\dfrac{1}{N}\sum_{i=1}^{N}\sum_{j=1}^{H \times W}\sum_{c=1}^{C} \hat{y}_{ab}^{u}[i,j,c]\log f_{\theta}(x_{ab}^{u})[i,j,c]
\label{eq: seg3}
\end{equation}
where $x_{ab}^{u}$ is the mixed image of $x_{a}^{u}$ and $x_{b}^{u}$ using $M$, and $\hat{y}_{ab}^{u}$ is the mixed pseudo label of $g_{\phi}(x_{a}^{u})$ and $g_{\phi}(x_{b}^{u})$ based on $M$.

\begin{table*}
\caption{Evaluation results for GTA5$\rightarrow$Cityscapes. The mIoU scores are calculated over $19$ classes of the Cityscapes validation set with respect to $0$, $100$, $200$, $500$, and $1000$ labeled samples. Note that the results of full labeled (2975) samples are removed since it violates the setting of our method which involves unlabeled target domain data during training. The symbol $-$ indicates evaluations that cannot be performed.
The best results are highlighted.}
\begin{center}
\begin{tabular}{l|l|ccccc}
  \hline
  Category & Method & $0$ & $100$ & $200$  & $500$ & $1000$ \\
  \hline
  		&AdaptSeg~\cite{tsai2018learning}         &42.4  &- &- &- &-\\
  		&Advent~\cite{vu2019advent}      &44.8  &- &- &- &-\\
  UDA&LTI~\cite{kim2020learning} &50.2  &- &- &- &- \\
		&PIT~\cite{lv2020cross} &50.6  &- &- &- &-\\
		&ProDA~\cite{zhang2021prototypical} &57.5  &- &- &- &- \\ 
  \hline
  Supervised&DeeplabV2 &- &41.9 &47.7 &55.5 &58.6 \\  
  \hline  
  		&CutMix~\cite{french2019semi}      &-  &50.8 &54.8 &61.7 &64.8\\
  SSL&DST-CBC~\cite{feng2020semi}      &-  &48.7 &54.1 &60.6 &63.2\\
  		&ClassMix~\cite{olsson2021classmix}      &-  &54.1 &61.4 &63.6 &66.3\\
		&ComplexMix~\cite{chen2021mask} &-  &53.9 &62.3 &64.1 &66.8\\
  \hline
  		& Baseline     &- &52.6 &53.6 &58.4 &61.6\\
  		& MME~\cite{saito2019semi}  &-  &52.6 &54.4 &57.6 &61.0\\
  SSDA&  ASS~\cite{wang2020alleviating} &-  &54.2 &56.0 &60.2&64.5\\
  		&  DLDM~\cite{chen2021semi} &- &\textbf{61.2} &60.5 &64.3 &66.6\\
  		&  Ours$^{1}$ (ClassMix) &-  &57.5 &62.8 &\textbf{65.9}&\textbf{66.9}\\
  		&  Ours$^{2}$ (ComplexMix) &-  &60.1 &\textbf{62.9} &65.7&66.8\\
  \hline
\end{tabular}
\label{table:gta}
\end{center}
\end{table*}

\section{Experimental Evaluation}

\medskip\noindent
\textbf{Data} 

\medskip\noindent
We validate the effectiveness of our proposed approach on two challenging synthetic-to-real SSDA semantic segmentation tasks. We use commonly used synthetic datasets (GTA5~\cite{richter2016playing} and SYNTHIA~\cite{ros2016synthia}) for the source domain, and a real-world image dataset (Cityscapes~\cite{cordts2016cityscapes}) for the target domain. The split into training and testing sets is done in a standard way as defined by other work (e.g. Chen \etal~\cite{chen2021semi}).
We train a model using a large amount of labeled source data, a small set of labeled target data, and the remaining unlabeled target data. 
\medskip\noindent
\textbf{Implementation details} 

\medskip\noindent
In our experiments 
we employ a Deeplab-V2~\cite{chen2017deeplab} model with a ResNet-101~\cite{he2016deep} backbone pretrained on ImageNet~\cite{deng2009imagenet}, as the base semantic segmentation (student) network $f_{\theta}$. 
We use the Pytorch deep learning framework to implement our approach on two NVIDIA-SMI GPUs with $16$ GB memory in total. The batch size for the training is set to $1$ and $4$ for the source and target domains respectively. For the target domain we use $4$ samples in each batch where $2$ are labeled images and $2$ are unlabeled. We did not attempt to tune this batch size parameter. 
During training, Stochastic Gradient Descent (SGD) is employed as the optimizer with momentum of $0.9$. We use a polynomial decay schedule~\cite{chen2017deeplab}.
The initial learning rate and weight decay are set to $5 \times 10^{-4}$ and $2.5 \times 10^{-4}$ respectively. The weight $\lambda$ in Eq.~\ref{eq: loss} is set to $1$.

\begin{figure*}[h]

\begin{minipage}[b]{1.0\linewidth}
  \centering
  \centerline{\includegraphics[scale=0.35]{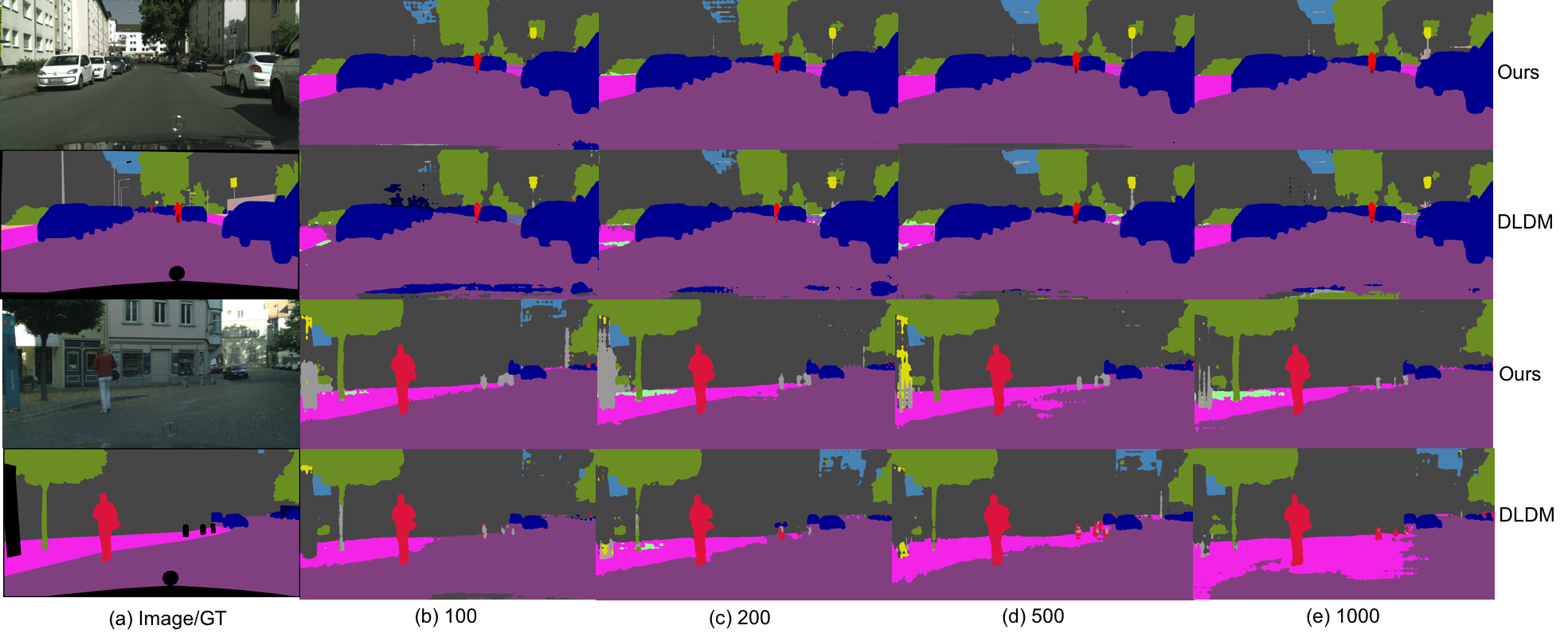}} 
\end{minipage}
\caption{GTA5$\rightarrow$Cityscapes qualitative semantic segmentation results comparing the proposed approach to DLDM~\cite{chen2021semi} using various amounts of labeled target data $(100, 200, 500, 1000)$. (a) Cityscapes validation examples and their associated ground truth (GT), (b)-(e) predicted segmentation maps using different numbers of labeled target images.}
\label{fig:eval_com_viz}
\end{figure*}

\medskip\noindent
\textbf{Results} 

\medskip\noindent
The proposed SSDDA approach is evaluated in the context of semantic segmentation using the standard mean intersection-over-union (mIoU) score computed on a separate Cityscapes validation set. We compare the proposed approach to state-of-the-art approaches and conduct an ablation study. Experimental results show that the proposed approach is capable of producing results which surpass the sate-of-the-art.

\medskip\noindent
\textbf{Compared Approaches} 

\medskip\noindent
Following the evaluation setting of Chen \etal~\cite{chen2021semi}, the baseline methods for comparison are divided into three main categories: UDA, SSL, and SSDA, where we add additional state-of-the-art methods to our comparison. UDA uses fully labeled source domain and unlabeled target domain data for training; SSL uses a small amount of labeled target and the remaining unlabeled target data for training; SSDA uses fully labeled source data, a small amount of labeled target data as well as reminding unlabeled target data to train the model. The supervised approach uses all available (source and target) labeled data without attempting to utilize unlabeled target data. All the trained models are evaluated on the same validation set. We evaluate two variants of our proposed approach differing by the way intra-domain adaptation is performed: Ours$^{1}$ uses ClassMix~\cite{french2019semi} whereas Ours$^{2}$ uses ComplexMix~\cite{chen2021mask}.

\medskip\noindent
\textbf{SYNTHIA$\rightarrow$Cityscapes Evaluation} 

\medskip\noindent
We evaluate the compared approaches using SYNTHIA as a source domain and Cityscapes as a target domain. We follow the evaluation setting of Chen \etal~\cite{chen2021semi} reporting the mIoU score of the $13$ classes shared between SYNTHIA and Cityscapes in Table~\ref{table:syn}. The columns of the table show the number of labeled target examples used for domain adaptation (out of approximately 3000 target domain examples). UDA approaches do not use labeled target data and so do not contain results for columns 100--1000. Likewise, the remaining approaches require labeled target data and so do not contain results for the 0 column. Note that the results of full labeled (2975) samples are removed since it violates the setting of our method which involves unlabeled target domain data during training.

As can be observed in the table: 
(1) our proposed models trained on different ratios of labeled target data outperform all existing methods within all three categories of UDA, SSL and SSDA;
(2) With $100$ labeled target examples the proposed approach outperforms all state-of-the-art UDA methods thus demonstrating the importance of semi-supervision.  
(3) With $200$ labeled target examples the proposed approach significantly narrows the performance gap with a fully supervised DeepLabV2 approach, thus demonstrating that the proposed approach is able to effectively utilize information from unlabeled target examples.
%
(4) With any amount of labeled target examples the proposed approach outperforms all SSL approaches thus demonstrating it can effectively utilize data from a different domain.
%
%
(5) The proposed approach outperforms at all mixing levels the state-of-the-art SSDA approach of Chen \etal~\cite{chen2021semi} (DLDM) which does not take into account intra-domain adaptation within the target domain, thus demonstrating the advantage of the dual domain adaptation in the proposed approach.
(6) The proposed method "Ours$^{2}$" using ComplexMix performs better than the "Ours$^{1}$" method using ClassMix for lower labeled data ratios thus demonstrating the importance of more complex mixing when the amounts of labeled target data are small.
%


\medskip\noindent
\textbf{GTA5$\rightarrow$Cityscapes Evaluation}

\medskip\noindent
To further validate the effectiveness of our proposed approach, we conduct an additional evaluation using GTA5 as a source domain and Cityscapes as a target domain. We follow the evaluation setting of Chen \etal~\cite{chen2021semi} reporting the mIoU score of the $19$ classes shared between SYNTHIA and Cityscapes. The evaluation results are presented in Table~\ref{table:gta}.
As shown in the table, our proposed approach outperforms all other methods for all ratios of labeled target domain examples. Hence, it further supports our arguments drawn in SYNTHIA$\rightarrow$Cityscapes.

Qualitative segmentation results are shown in Figure~\ref{fig:eval_com_viz} comparing DLDM~\cite{chen2021semi} to the proposed SSDDA approach using various amounts of labeled target examples ($100$, $200$, $500$ and $1000$). As can be observed, the results obtained by proposed approach resemble better the ground truth for larger amounts of labeled images.

\medskip\noindent
\textbf{Ablation Study}

\medskip\noindent
To further demonstrate the effectiveness of our approach, we conduct an additional ablation study. Our approach is comprised of two domain adaptation strategies and in the ablation study we demonstrate the contribution in using dual (simultaneous) domain adaptation to using the single domain adaptation components: cross-domain-only and intra-domain-only. The evaluation is performed when using GTA5 as the source domain and Cityscapes as the target domain. Quantitative evaluation results are shown in Table~\ref{table:abla}. As can be observed in the table, intra-domain-only adaptation obtains better results compared with cross-domain-only adaptation, while our proposed dual-domain adaptation approach achieves the best results. Overall, this demonstrates that our dual-domain adaptation approach successfully transfers information from the source domain to the target domain while effectively alleviating intra-domain discrepancy generated by the cross-domain adaptation process. 

\begin{table}
\caption{Ablation study results for cross-domain-only and intra-domain-only for GTA5$\rightarrow$Cityscapes. The mIoU score are calculated over $19$ classes of Cityscapes validation set with  $100$, $200$, $500$, and $1000$ labeled examples.}
\begin{center}
\begin{tabular}{l|cccc}
  \hline
  Method & $100$ & $200$  & $500$ & $1000$ \\
  \hline
  		Cross-domain only        &47.9  &50.2 &58.4 &60.0\\
  		Intra-domain only      &53.9  &62.2 &64.1 &66.7\\
  	Ours$^{1}$ (ClassMix) &\textbf{57.5} &\textbf{62.8} &\textbf{65.9}&\textbf{66.9}\\
  \hline
\end{tabular}
\label{table:abla}
\end{center}
\end{table}

\section{Conclusion}
In this paper we propose a new semi-supervised dual-domain adaptation (SSDDA) approach for semantic segmentation. Cross-domain adaptation is achieved by a student network, while a separate student-teacher network sharing weights with the former student network is used to address intra-domain shift. We validate the effectiveness of the proposed approach and show that by utilizing useful information from the source and target domain data, we can surpass state-of-the-art performance when evaluated on two standard synthetic-to-real tasks.

\bibliography{egbib}
\end{document}